%% file: sample-sigconf.tex
\documentclass[sigconf]{acmart}

\AtBeginDocument{%
  \providecommand\BibTeX{{%
    \normalfont B\kern-0.5em{\scshape i\kern-0.25em b}\kern-0.8em\TeX}}}

\usepackage{CJK}
\usepackage{multirow}
\usepackage{amsmath}

\usepackage{amssymb}

\copyrightyear{2021}
\acmYear{2021}
\setcopyright{acmcopyright}\acmConference[MM '21]{Proceedings of the 29th ACM International Conference on Multimedia}{October 20--24, 2021}{Virtual Event, China}
\acmBooktitle{Proceedings of the 29th ACM International Conference on Multimedia (MM '21), October 20--24, 2021, Virtual Event, China}
\acmPrice{15.00}
\acmDOI{10.1145/3474085.3475442}
\acmISBN{978-1-4503-8651-7/21/10}

\begin{document}
\fancyhead{}




\title{Neural Free-Viewpoint Performance Rendering under Complex Human-object Interactions}

\input{./sections/Authors.tex}


\begin{abstract}
    \input{./sections/Abstract.tex}
\end{abstract}



\begin{teaserfigure}
  \centering
  \vspace{-5pt}
  \includegraphics[width=0.95\textwidth]{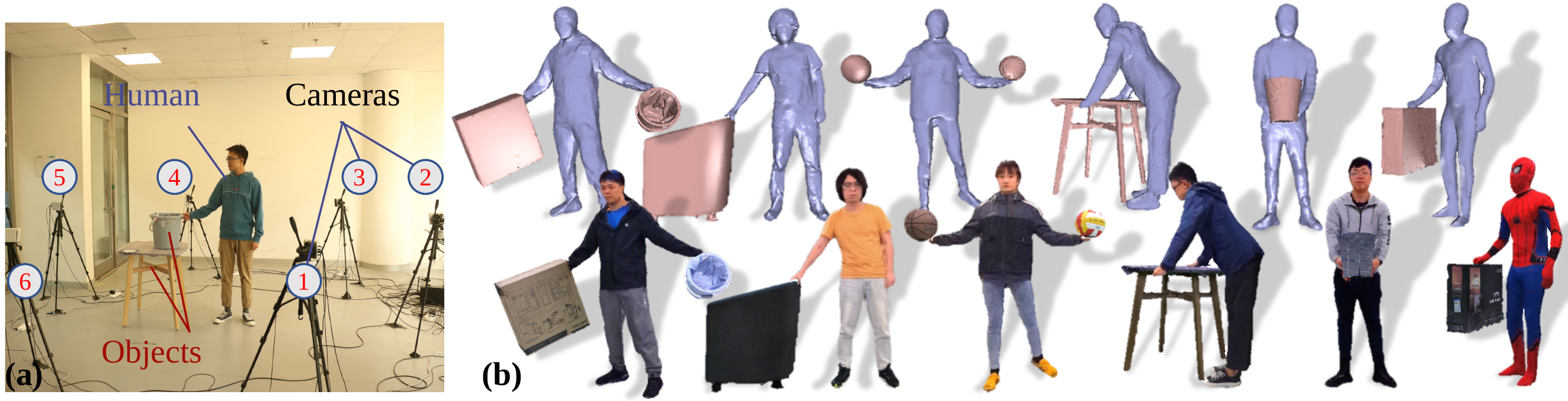}
  \vspace{-10pt}
  \caption{Our approach achieves photo-realistic reconstruction results of human activities in novel views under challenging human-object interactions, using only six RGB cameras. (a) Capture setting. (b) Our results.}
  \label{fig:teaser}
\end{teaserfigure}

\maketitle


\begin{CJK}{UTF8}{gbsn}
    \input{sections/Intro}

\input{sections/Related}
\input{sections/Overview}

    \input{sections/Method}
    \input{sections/Experiment}

    \input{sections/Conclusion}

    \input{sections/Acknowledgments}
\end{CJK}

\bibliographystyle{ACM-Reference-Format}
\balance
\bibliography{reference}

\newpage
\nobalance
\appendix
\section{More Implementation Details}
We normalize the camera system to $m$ unit and transform it to align the system center with the coordinate origin. Our voxel encoder is a forward 3D CNN like ~\cite{chibane2020implicit}, outputs 1, 16, 32, 32, 32 dimension features from each layer. Our image encoder outputs 128 dimension feature in the final layer. A 1-d depth feature is produced by perspective projection like ~\cite{PIFU_2019ICCV}. We concatenate these features and pass them through a MLP like ~\cite{PIFU_2019ICCV} with dimensions of 242, 1024, 512, 256, 128, 1. For the training data of human reconstruction, we sample points 2 $cm$, 3 $cm$, 6 $cm$ and uniformly in the 3D space. We group the surface samplings, and combine it with uniform sampling using a ratio of 16 : 1. We train the human reconstruction network with Adam optimizer at the batch size of 4. The learning rate is 1e-4 and sampling strategy is (3 $cm$,6 $cm$, uniform) for the first 20 epochs. Then we 
reduce learning rate to 1e-5 and change sampling strategy to (2 $cm$,3 $cm$, uniform) for further 100 epochs. In the object tracking stage, we use silhouette rendering in pytorch3d~\cite{ravi2020accelerating} under perspective projection. The Adam optimizer with learning rate of 1e-2 is used for a 100-iteration tracking optimization. For the neural blending network, we train it with Adam optimizer at the learning rate of 1e-4 and the batch size of 4. We train this network 44,000 iterations.

\section{Camera setting}
Pioneer work ~\cite{NeuralHumanFVV2021CVPR} evaluates the influence of sparse view number in their section 5, and finds that six cameras serve as a good compromise between rendering artifacts and the camera number. Since human-object interaction is even more challenging, we choose ``six'' for stable object tracking, robust human tracking and high-fidelity rendering. Finally, we place the cameras uniformly to minimize the average artifacts in circle rendering.

\section{Efficiency}
Compared with SOTAs, our method doesn't need per-scene training which improves efficiency in practice greatly(save several hours per-scene). However, one promising direction is to add the real-time ability into our framework to enable immersive telepresence under the human-object interaction scenario. Our method takes 0.2s to estimate SMPL, 7s to generate mesh, 7s to track object, and 0.25s to render in single frame. We believe the coarse-to-fine strategy~\cite{NeuralHumanFVV2021CVPR} and the end-to-end 6DoF estimation~\cite{peng2019pvnet} can accelerate human reconstruction and object tracking respectively.

\end{document}

%% file: sections/Authors.tex
\newcommand{\tsc}[1]{\textsuperscript{#1}} 

\author{
  \hspace{0.1em} Guoxing Sun\tsc{1},
  \hspace{0.1em} Xin Chen\tsc{1,2,3},
  \hspace{0.1em} Yizhang Chen\tsc{1},
  \hspace{0.1em} Anqi Pang\tsc{1,2,3},
  \hspace{0.1em} Pei Lin\tsc{1},
  \hspace{0.1em} Yuheng Jiang\tsc{1},\\
  Lan Xu\tsc{1}, \hspace{0.1em} Jingya Wang\tsc{1\footnotemark[2]}, \hspace{0.1em} Jingyi Yu\tsc{1\footnotemark[2]}
      }

\affiliation{%
 \institution{\tsc{1}ShanghaiTech University, School of Information Science and Technology, Shanghai Engineering               Research Center of Intelligent Vision and Imaging\country{China}
 \\
            \tsc{2}Shanghai Institute of Microsystem and Information Technology, Chinese Academy of Sciences\country{China} \\
            \tsc{3}University of Chinese Academy of Sciences\country{China} 
            }
  \institution{\{sungx,
  \hspace{0.2em}chenxin2,
  \hspace{0.2em}chenyzh,
  \hspace{0.2em}pangaq,
  \hspace{0.2em}linpei,
  \hspace{0.2em}jiangyh2,
  \hspace{0.2em}xulan1,
  \hspace{0.2em}wangjingya,
  \hspace{0.2em}yujingyi
  \} @shanghaitech.edu.cn}
}

%% file: sections/Abstract.tex
4D reconstruction of human-object interaction is critical for immersive VR/AR experience and human activity understanding.
Recent advances still fail to recover fine geometry and texture results from sparse RGB inputs, especially under challenging human-object interactions scenarios. 
In this paper, we propose a neural human performance capture and rendering system to generate both high-quality geometry and photo-realistic texture of both human and objects under challenging interaction scenarios in arbitrary novel views, from only sparse RGB streams. 
To deal with complex occlusions raised by human-object interactions, we adopt a layer-wise scene decoupling strategy and perform volumetric reconstruction and neural rendering of the human and object.
Specifically, for geometry reconstruction, we propose an interaction-aware human-object capture scheme that jointly considers the human reconstruction and object reconstruction with their correlations. Occlusion-aware human reconstruction and robust human-aware object tracking are proposed for consistent 4D human-object dynamic reconstruction. 
For neural texture rendering, we propose a layer-wise human-object rendering scheme, which combines direction-aware neural blending weight learning and spatial-temporal texture completion to provide high-resolution and photo-realistic texture results in the occluded scenarios.
Extensive experiments demonstrate the effectiveness of our approach to achieve high-quality geometry and texture reconstruction in free viewpoints for challenging human-object interactions.

%% file: sections/Intro.tex
\section{Introduction}

The rise of virtual and augmented reality (VR and AR) has increased the
the demand of the human-centric 4D (3D spatial plus 1D time) content generation, with numerous applications from entertainment to commerce, from gaming to education.
Further reconstructing the 4D models and providing photo-realistic rendering of challenging human-object interaction scenarios conveniently evolves as a cutting-edge yet bottleneck technique.

Early high-end solutions~\cite{motion2fusion,TotalCapture,collet2015high,UnstructureLan} rely on multi-view dome-based setup to achieve high-fidelity reconstruction and rendering, which are expensive and difficult to be deployed.
%
The recent low-end volumetric approaches~\cite{UnstructureLan,robustfusion,su2021robustfusion,DoubleFusion,BodyFusion,tao2021function4d} have enabled light-weight performance reconstruction by leveraging the RGBD sensors and modern GPUs.
However, they still suffer from unrealistic texturing results and heavily rely on depth cameras which are not as ubiquitous as color cameras.
The recent neural rendering techniques~\cite{Wu_2020_CVPR,NeuralVolumes,nerf} bring huge potential for human reconstruction and photo-realistic rendering from only RGB inputs.
In particular, the approaches with implicit function~\cite{saito2019pifu,saito2020pifuhd,zheng2020pamir} reconstruct clothed humans with fine geometry details but are restricted to only human without modeling human-object interactions.
For photo-realistic human performance rendering, various data representation have been explored, such as point-clouds~\cite{Wu_2020_CVPR,pang2021fewshot}, voxels~\cite{NeuralVolumes}, implicit representations~\cite{nerf,peng2021neural,pumarola2020d,tretschk2020nonrigid} or hybrid neural texturing~\cite{NeuralHumanFVV2021CVPR}. 
However, existing solutions rely on doom-level dense RGB sensors or are limited to human priors without considering the joint rendering of human-object interactions.
Besides, various researchers~\cite{PSI2019,zhang2020object,2020phosa_Arrangements,PLACE:3DV:2020,HPS,Hassan:CVPR:2021,PatelCVPR2021,GRAB:2020} models the interaction between humans and the objects or the surrounding environments.
However, they only recover the naked human bodies or a visually plausible but not accurate spatial arrangement.
Few researchers explore to combine volumetric modeling and photo-realistic novel view synthesis under the human-object interaction scenarios, especially for the light-weight sparse RGB setting.

In this paper, we attack these challenges and present a human-object neural volumetric rendering using only sparse RGB cameras surrounding the performer and objects.
As illustrated in Fig.~\ref{fig:teaser}, our novel approach generates both high-quality geometry
and photo-realistic texture of human activities in novel views for both the performers and objects under challenging interaction scenarios, whilst maintaining the light-weight setup.

Generating such a free-viewpoint video and achieving robust volumetric reconstruction of human activities under challenging human-object interactions is non-trivial.
Our key idea is to embrace a layer-wise scene decoupling strategy and perform volumetric reconstruction and rendering of the target human and object
separately, so as to model the challenging occlusion explicitly for human-object interactions. 
To this end, we first apply an off-the-shelf instance segmentation approach to the six RGB input streams to distinguish the human and object separately.
Then, for robust geometry reconstruction, we introduce a novel implicit human-object capture scheme to model the mutual influence between human and object.
For the human reconstruction, we perform a neural implicit geometry generation to jointly utilize both the pixel-aligned image features and global human motion priors with the aid of an occlusion-aware training data augmentation. 
For the objects, we perform a template-based object alignment for the first frame and human-aware tracking to maintain temporal consistency and prevent the segmentation uncertainty caused by interaction.
Finally, for photo-realistic rendering, based on the geometry proxy above, a layer-wise human-object rendering scheme is proposed to disentangle the human and object separately.
Specifically, we adopt template-based texturing with color correction for the object and extend the neural texturing scheme~\cite{NeuralHumanFVV2021CVPR} into our interaction scenarios with severe human-object occlusion.
We propose a direction-aware neural texturing blending scheme that encodes the occlusion information explicitly, and adopts a spatial-temporal texture completion for the occluded regions based on the human motion priors.
With the aid of such occlusion analysis, our texturing scheme maps the input adjacent images into a photo-realistic texture output of human-object activities in the target view through efficient blending weight learning, without requiring further per-scene training.
To summarize, our main technical contributions include:
\begin{itemize}
	\item We present a neural human performance rendering scheme for challenging human-object interaction scenarios using only sparse RGB cameras, which can reconstruct high-quality texture and geometry results of human activities, achieving significant superiority to existing state-of-the-art.
	
	\item We propose an interaction-aware human-object capture sch-eme that combines occlusion-aware neural implicit human geometry generation and robust human-aware object tracking for consistent 4D human-object dynamic reconstruction.
	
	\item We introduce a layer-wise human-object rendering scheme, which utilizes occlusion-aware neural blending weight learning and spatial-temporal texture completion to provide high-resolution and photo-realistic texture results.
\end{itemize}

%% file: sections/Related.tex
\section{Related Work}

  \begin{figure*}[t]
    \centering
    \includegraphics[width=\linewidth]{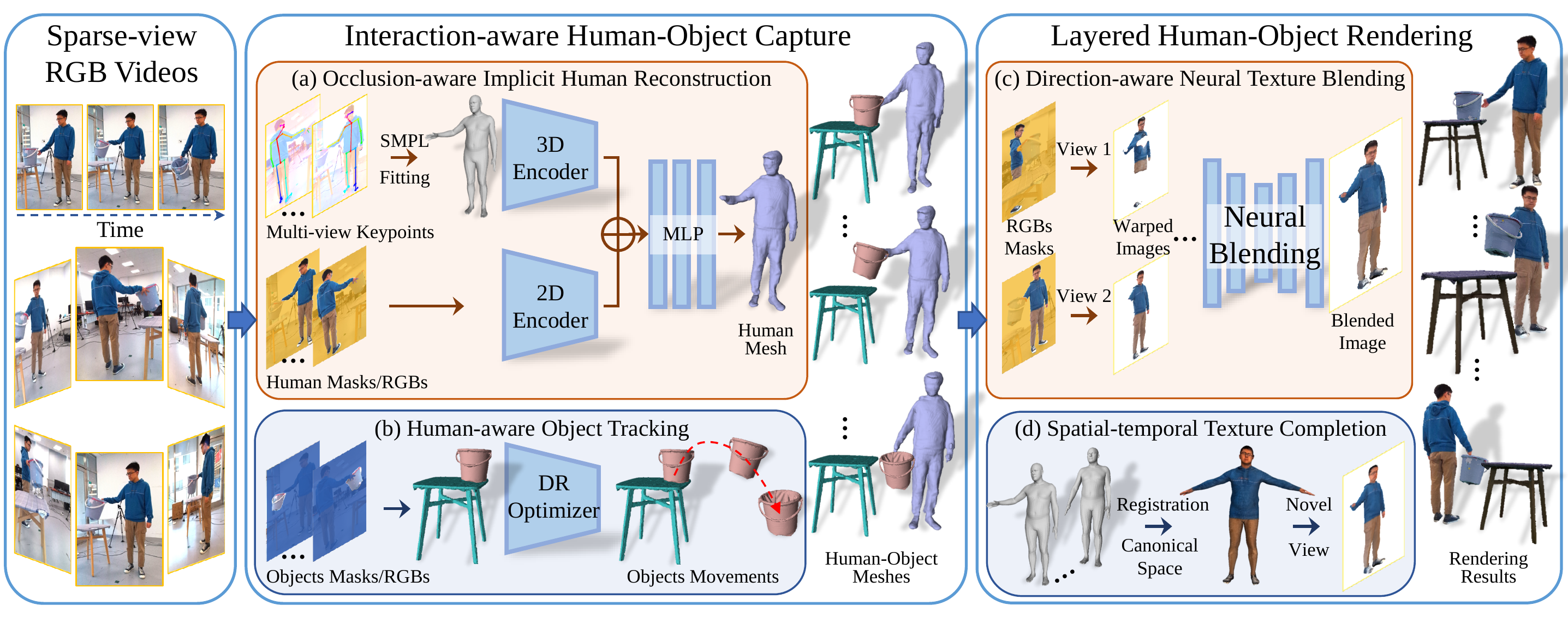}
    \vspace{-20pt}
    \caption{The overview of our approach. Given the six RGB stream inputs surrounding the performer and objects, our approach generates high-quality human-object meshes and free-view rendering results. ``DR'' indicates differentiable rendering.
    }
    \vspace{-10pt}
    \label{fig:pipeline}
  \end{figure*}

\noindent{\textbf{Human Performance Capture.}}
Markerless human performance capture techniques have been widely investigated to achieve human free-viewpoint video or reconstruct the geometry. 
The high-end solutions~\cite{motion2fusion,TotalCapture,collet2015high,chen2019tightcap} adopt studio-setup with dense cameras to produce high-quality reconstruction and surface motion, but the synchronized and calibrated multi-camera systems are both difficult to deploy and expensive.
The recent low-end approaches~\cite{Xiang_2019_CVPR,LiveCap2019tog,chen2021sportscap, he2021challencap} enable light-weight performance capture under the single-view setup or even hand-held capture setup or drone-based capture setup~\cite{xu2017flycap}.
However, these methods require a naked human model or pre-scanned template. 
Volumetric fusion based methods~\cite{newcombe2015CVPR,DoubleFusion,BodyFusion,HybridFusion} enables free-form dynamic reconstruction. But they still suffer from careful and orchestrated motions, especially for a self-scanning process where the performer turns around carefully to obtain complete reconstruction. 
\cite{robustfusion} breaks self-scanning constraint by introducing implicit occupancy method.
 All these methods suffer from the limited mesh resolution leading to uncanny texturing output. Recent method~\cite{mustafa2020temporally} leverages unsupervised temporally coherent human reconstruction to generate free-viewpoint rendering. It is still hard for this method to get photo-realistic rendering results.
Comparably, our approach enables the high-fidelity capture of human-object interactions and eliminates the additional motion constraint under the sparse view RGB camera settings.

\noindent{\textbf{Neural Rendering.}}
The recent progress of differentiable neural rendering brings huge potential for 3D scene modeling and photo-realistic novel view synthesis. Researchers explore various data representations to pursue better performance and characteristics, such as point-clouds~\cite{Wu_2020_CVPR,aliev2019neural,suo2020neural3d}, voxels~\cite{lombardi2019neural}, texture meshes~\cite{thies2019deferred,liu2019neural} or implicit functions~\cite{park2019deepsdf,nerf,meng2021gnerf,chen2021mvsnerf,wang2021mirrornerf,luo2021convolutional}. 
However, these methods require inevitable pre-scene training to a new scene.
For neural modeling and rendering of dynamic scenes, NHR~\cite{Wu_2020_CVPR} embeds spatial features into sparse dynamic point-clouds, Neural Volumes~\cite{NeuralVolumes} transforms input images into a 3D volume representation by a VAE network.
More recently, \cite{park2020deformable,pumarola2020d,li2020neural,xian2020space,tretschk2020non,peng2021neural,zhang2021editable} extend neural radiance field (NeRF)~\cite{nerf} into the dynamic setting. 
They learn a spatial mapping from the canonical scene to the current scene at each time step and regress the canonical radiance field. 
However, for all the dynamic approaches above, dense spatial views or full temporal frames are required in training for high fidelity novel view rendering, leading to deployment difficulty and unacceptable training time overhead. Recent approaches~\cite{peng2021neural} and ~\cite{NeuralHumanFVV2021CVPR} adopt a sparse set of camera views to synthesize photo-realistic novel views of a performer. However, in the scenario of human-object interaction, these methods fail to generate both realistic performers and realistic objects.
Comparably, our approach explores the sparse capture setup and fast generates photo-realistic texture of challenging human-object interaction in novel views.

\noindent{\textbf{Human-object capture.}}
Early high-end work~\cite{collet2015high} captures both human and objects by reconstruction and rendering with dense cameras. 
Recently, several works explore the relation between human and scene to estimate 3D human pose and locate human position~\cite{hassan2019resolving,HPS,liu20204d}, naturally place human~\cite{PSI2019,PLACE:3DV:2020,hassan2021populating} or predict human motion~\cite{cao2020long}. 
Another related direction~\cite{GRAB:2020,hampali2021handsformer,liu2021semi} models the relationship between hand and objects for generation or capture.
PHOSA~\cite{2020phosa_Arrangements} runs human-object capture without any scene- or object-level 3D supervision using constraints to resolve ambiguity. 
However, they only recover the naked human bodies and produce a visually reasonable spatial arrangement.
A concurrent close work is RobustFusion(journal)~\cite{su2021robustfusion}. They capture human and objects by volumetric fusion respectively, and track object by Iterative Closest Point (ICP). 
However, their texturing quality is limited by mesh resolution and color representation, and the occluded region is ambiguous in 3D space.
Comparably, our approach enables photo-realistic novel view synthesis and accurate human object arrangement in 3D world space
under the human-object interaction for the light-weight sparse RGB settings.

%% file: sections/Overview.tex
\section{Overview}
An overview of the proposed architecture is depicted in Fig. \ref{fig:pipeline}. Given the sparse-view RGB video inputs, we introduce a coarse-to-fine multi-stage neural human-object rendering scheme to handle challenging scenarios with severe occlusions and multi objects. Our approach captures high-fidelity human-object geometry and arrangement by interaction-aware human-object capture and generates photo-realistic novel view rendering results by layered human-object rendering. 
A brief introduction of our main components is provided as follows:

\noindent{\textbf{Occlusion-aware Implicit Human Reconstruction (Fig.~\ref{fig:pipeline} (a)).}}
We perform a neural implicit geometry generation to utilize both the pixel-aligned image features and global human motion priors for the human reconstruction. Through the occlusion-aware human reconstruction scheme, our approach reconstructs high-fidelity human geometry under different occlusion scenarios.

\noindent{\textbf{Human-aware Object Tracking (Fig. \ref{fig:pipeline} (b)).}}
For the objects interacted with human, we perform a template-based object alignment and human-aware object tracking to maintain temporal consistency.
Efficient differentiable renderers are used to predict the 6DoF pose of the object by comparing rendered masks. 
Our continuous object tracking works robustly under the world space by jointly considering object masks, occlusion, and mesh-intersection. 

\noindent{\textbf{Direction-aware Neural Texture Blending (Fig. \ref{fig:pipeline} (c)).}}
For photo-realistic rendering, a layer-wise human-object rendering scheme is proposed to disentangle the human and object separately.
We adopt template-based texturing with color correction for the object and extend the neural texturing scheme~\cite{NeuralHumanFVV2021CVPR} into our interaction scenarios with severe human-object occlusion.
To deal with occlusion, we propose a direction-aware neural texturing blending scheme to explicitly encode the occlusion information and balance the quality of the warped images with the angle map between two source views and the novel view.

\noindent{\textbf{Spatial-temporal Texture Completion (Fig. \ref{fig:pipeline} (d)).}}
Based on the geometry from the results of the human-object capture, we adopt a spatial-temporal texture completion for the occluded regions based on the human motion priors.
We generate a texture-completed proxy in the canonical human space firstly. We thus use this information to complete the missing texture in a novel view.

We describe more details for each component in Sec.\ref{sec:method}

%% file: sections/Method.tex
\section{Method}\label{sec:method}

\subsection{Interaction-aware Human-Object Capture}\label{sec:human_capture}
Classical multi-view stereo reconstruction approaches \citep{Furukawa2013,Strecha2008,Newcombe2011,collet2015high} and recent neural rendering approaches \citep{Wu_2020_CVPR,NeuralVolumes,nerf} rely on multi-view dome based setup to achieve high-fidelity reconstruction and rendering results.
However, they suffer from both sparse-view inputs and occlusion of objects.
To this end, we propose a novel implicit human-object capture scheme to model the mutual influence between human and object from only sparse-view RGB inputs.

\noindent{\textbf{(a) Occlusion-aware Implicit Human Reconstruction.}}
For the human reconstruction, we perform a neural implicit geometry generation to jointly utilize both the pixel-aligned image features and global human motion priors with the aid of an occlusion-aware training data augmentation.

Without dense RGB cameras and depth cameras, traditional multi-view stereo approaches \citep{collet2015high,motion2fusion} and depth-fusion approaches \citep{KinectFusion,UnstructureLan,robustfusion} can hardly reconstruct high-quality human meshes.
With implicit function approaches \citep{PIFU_2019ICCV,PIFuHD}, we can generate fine-detailed human meshes with sparse-view RGB inputs.
However, the occlusion from human-object interaction can still cause severe artifacts.
To end this, we thus utilize the pixel-aligned image features and global human motion priors.

Specifically, we adopt the off-the-shelf instance segmentation approach \citep{Bolya_2019_ICCV} to obtain human and object masks, thus distinguishing the human and object separately from the sparse-view RGB input streams.
Meanwhile, we apply the parametric model estimation to provide human motion priors for our implicit human reconstruction.
We voxelized the mesh of this estimated human model to represent it with a volume field.

We give both the pixel-aligned image features and global human motion priors in volume representation to two different encoders of our implicit function, as shown in Fig. \ref{fig:pipeline} (a).
Different from \cite{2020phosa_Arrangements} with only a single RGB input, we use pixel-aligned image features from the multi-view inputs and concatenate them with our encoded voxel-aligned features.
We finally decode the pixel-aligned and voxel-aligned feature to occupancy values with a multilayer perceptron (MLP).

For each query 3D point $P$ on the volume grid, we follow PIFu \citep{saito2019pifu} to formulate the implicit function $f$ as:
\begin{align}
	f( \Phi(P),\Psi(P),Z(P)) & = \sigma : \sigma \in [0.0, 1.0],             \\
	\Phi(P)                  & = \frac{1}{n} \sum_{i}^{n}F_{I_{i}}(\pi_{i}(P)), \\
	\Psi(P)                  & =  G(F_{V},P),
\end{align}
where $p = \pi_{i}(P)$ denotes the projection of 3D point to camera view $i$, $F_{I_{i}}(x)= g(I_{i}(p))$ is the image feature at $p$.
$\Psi(P) = G(F_{V},P)$ denotes the voxel aligned features at $P$, $F_{V}$ is the voxel feature.
To better deal with occlusion, we introduce an occlusion-aware reconstruction loss to enhance the prediction at the occluded part of human.
It is formulated as:
	\begin{align}
		 & \mathcal{L}_{\sigma} = \lambda_{occ}\sum_{t=1}^T \left\| \sigma_{occ}^{gt} - \sigma_{occ}^{pred} \right\|_2^2 + \lambda_{vis}\sum_{t=1}^T \left\| \sigma_{vis}^{gt} - \sigma_{vis}^{pred} \right\|_2^2.
	\end{align}

Here, $\lambda_{occ}$ and $\lambda_{vis}$ represent the weight of occlusion points and visible points, respectively.
$\sigma_{occ}$ and $\sigma_{vis}$ are the training sampling points at the occlusion area and visible area.

\begin{figure}[t]
    \centering
    \includegraphics[width=\linewidth]{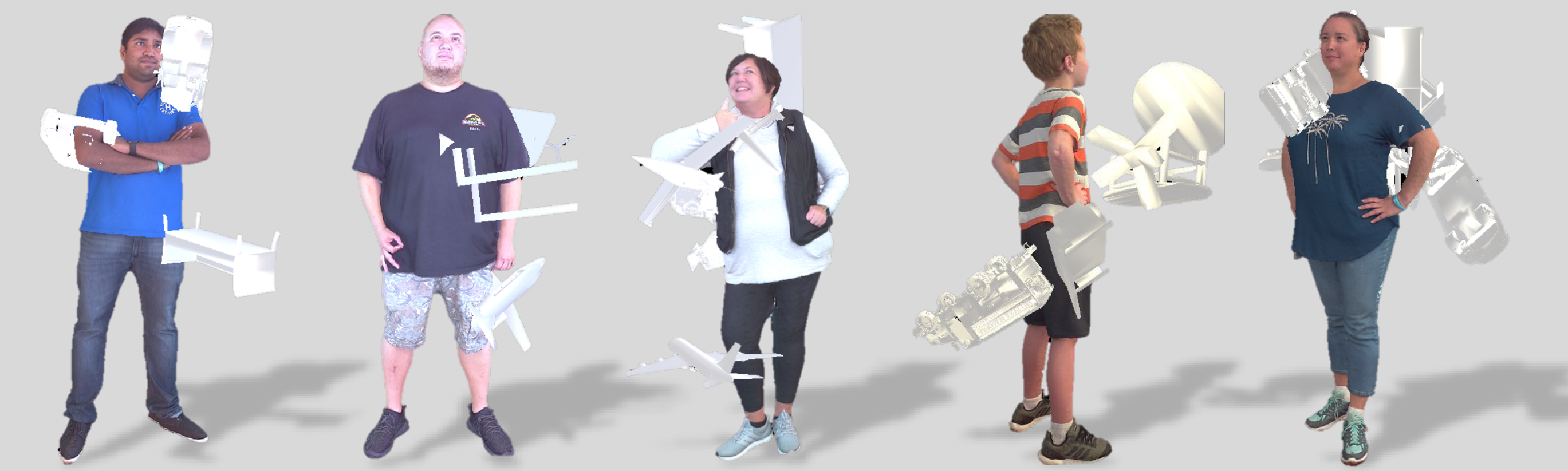}
    \vspace{-10pt}
    \caption{Illustration of our synthetic 3D data with both human and objects.}
    \vspace{-15pt}
    \label{fig:DataAugmentation}
\end{figure}

\begin{figure*}[t]
	\centering
	\includegraphics[width=\linewidth]{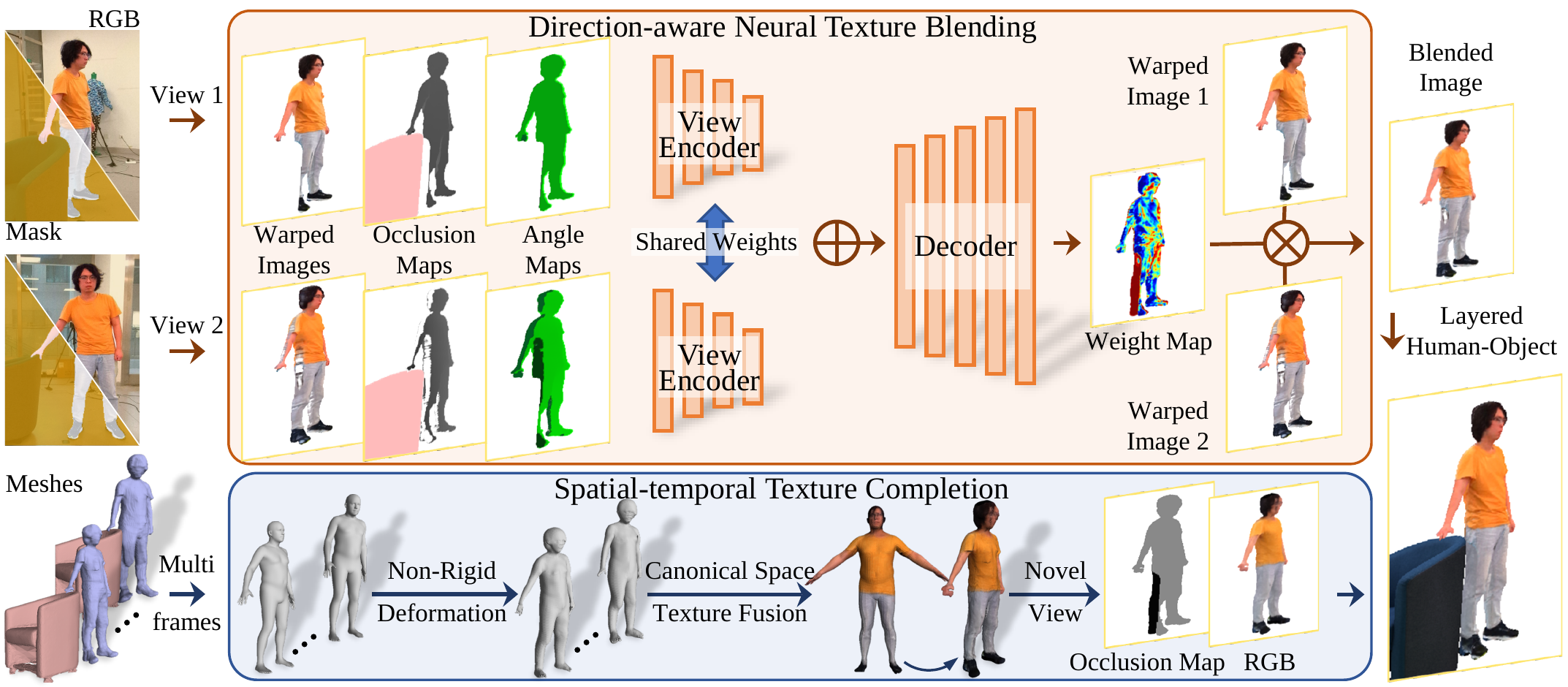}
	\caption{Illustration of our layered human-object rendering approach, which not only includes a direction-aware neural texture blending scheme to encode the occlusion information explicitly but also adopts a spatial-temporal texture completion for the occluded regions based on the human motion priors.}
	\vspace{-10pt}
	\label{fig:pipeline_net}
\end{figure*}

For the detail of the parametric model estimation, we fit the parametric human model, SMPL \citep{SMPL2015}, to capture occluded human with the predicted 2D keypoints.
Specifically, we use Openpose \citep{Openpose} as our joint detector to estimate 2D human keypoints from sparse-view RGB inputs.
To estimate the pose/shape parameters of SMPL as our human prior for occluded human, we formulate the energy function $\boldsymbol{E}_{\mathrm{prior}}$ of this optimization as:
\begin{align} \label{eq:opt}
	\boldsymbol{E}_{\mathrm{prior}}(\boldsymbol{\theta}_t, \boldsymbol{\beta}) = \boldsymbol{E}_{\mathrm{2D}} + \lambda_{\mathrm{T}}\boldsymbol{E}_{\mathrm{T}}
\end{align}
Here, $\boldsymbol{E}_{\mathrm{2D}}$ represents the re-projection constraint on 2D keypoints detected from sparse-view RGB inputs, while $\boldsymbol{E}_{\mathrm{T}}$ enforces the final pose and shape to be temporally smooth.
$\boldsymbol{\theta}_t$ is the pose parameters of frame $t$, while $\boldsymbol{\beta}$ is the shape parameters.
Note that this temporal smoothing enables globally consistent capture during the whole sequence, and benefits the parametric model estimation when some part of the body is gradually occluded.
We follow \cite{he2021challencap} to formulate the 2D term $\boldsymbol{E}_{\mathrm{2D}}$ and the temporal term $\boldsymbol{E}_{\mathrm{T}}$ under the sparse-view setting.

Moreover, we apply an occlusion-aware data augmentation to reduce the domain gap between our training set and the challenging human-object interaction testing set.
Specially, we randomly sample some objects from ShapeNet dataset~\cite{chang2015shapenet}.
We then randomly rotate and place them around human before training, as shown in Fig. \ref{fig:DataAugmentation}.
By simulating the occlusion of human-object interaction, our network is more robust to occluded human features.

With both the pixel-aligned image features and the statistical human motion priors under this occlusion-aware data augmentation training, our implicit function generates high-quality human meshes with only spare RGB inputs and occlusions from human-object interaction.

\noindent{\textbf{(b) Human-aware Object Tracking.}}
For the objects around the human, people recover them from depth maps~\cite{new2011kinect}, implicit fields~\cite{mescheder2019occupancy}, or semantic parts~\cite{chen2018autosweep}. We perform a template-based object alignment for the first frame and human-aware tracking to maintain temporal consistency and prevent the segmentation uncertainty caused by interaction. With the inspiration of PHOSA \cite{2020phosa_Arrangements}, we consider each object as a rigid body mesh.

To faithfully and robustly capture object in 3D space as time going, we introduce a human-aware tracking method.
Expressly, we assume objects are rigid bodies and transforming rigidly in the human-object interaction activities.
So the object mesh $O_{t}$ at frame $t$ can be represented as: $O_{t} = R_{t}O_{t-1}+T_{t}$.
Based on the soft rasterization rendering~\cite{ravi2020accelerating}, the rotation $R_{t}$ and the translation $T_{t}$ can be naively optimized by comparing $\mathcal{L}_{2}$ norm between the rendered silhouette $S_{t}^{i}$ and object mask $\mathcal{M}o_{t}^{i}$.

Human is also an important cue to locate the object position.
From the 2D perspective, when objects are occluded by the human at a camera view, the  $\mathcal{L}_{2}$ loss between rendered silhouette and occluded mask will lead to the wrong object location due to the wrong guidelines at the occluded area.
So we remove the occluded area affected by human mask $\mathcal{M}h_{t}^{i}$ when computing the $\mathcal{L}_{2}$ loss.
From the 3D perspective, human can not interpenetrate an rigid object, so we also add an interpenetration loss $\mathcal{L}_{P}$~\cite{jiang2020mpshape} to regularize optimization. Our total object tracking loss is:
\begin{align}
	\mathcal{L}_{track} = \lambda_{1}\sum_{i=0}^{n}\| \mathcal{B}(\mathcal{M}h_{t}^{i}==0) \odot  S_{t}^{i} - \mathcal{M}o_{t}^{i}  \|  + \lambda_{2}\mathcal{L}_{P},
\end{align}
where n denotes view numbers, $\lambda_{1}$ denotes weight of silhouette loss, $\lambda_{2}$ denotes weight of interpenetration loss, $\mathcal{B}$ represents an binary operation, it returns 0 when the condition is true, else 1.

Our implicit human-object capture utilizes both the pixel-aligned image features and global human motion priors with the aid of an occlusion-aware training data augmentation, and captures objects with the template-based alignment and the human-aware tracking to maintain temporal consistency and prevent the segmentation uncertainty caused by interaction. Thus, our approaches can generate high-quality human-object geometry with sparse inputs and occlusions.

\subsection{Layered Human-Object Rendering}\label{sec:rendering}
We introduce a neural human-object rendering pipeline to encode local fine-detailed human geometry and texture features from adjacent input views, so as to produce photo-realistic layered output in the target view, as illustrated in Fig. \ref{fig:pipeline_net}.

\begin{figure*}[t]
	\centering
	\includegraphics[width=\linewidth]{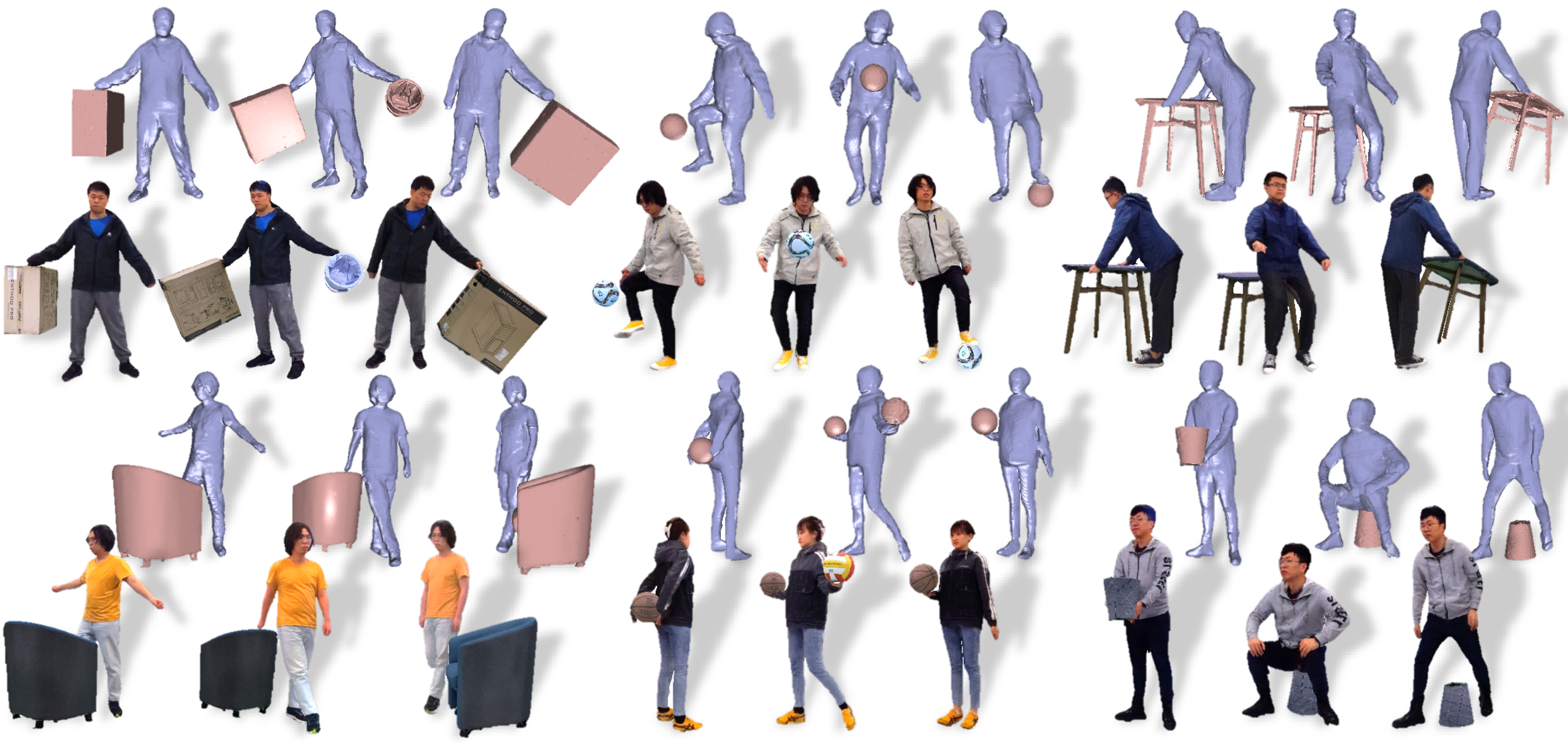}
	\vspace{-20pt}
	\caption{The geometry and texture results of our proposed approach, which generates photo-realistic rendering results and high fidelity geometry on a various of sequences, such as rolling a box, playing with balls.}
    \vspace{-10pt}
	\label{fig:gallery}
\end{figure*}

\noindent{\textbf{(c) Direction-aware Neural Texture Blending.}} \label{sec:neuralBlending}
While traditional image-based rendering approaches always show the artifacts with the sparse-view texture blending, we follow \cite{NeuralHumanFVV2021CVPR} to propose a direction-aware neural texture blending approach to render photo-realistic human in the novel view.
For a novel view image $I_{n}$, the linear combination of two source view $I_{1}$ and $I_{2}$ with blending weight map $W$ is formulate as:
\begin{align}
	I_{n} = W \cdot I_{1} + (1 - W) \cdot I_{2}.
\end{align}
However, in the sparse-view setting, the neural blending approach \citep{NeuralHumanFVV2021CVPR} can still generate unsmooth results. As the reason of these artifacts, the imbalance of angles between two source views with a novel view will lead to the imbalance wrapped image quality.

Different from \citet{NeuralHumanFVV2021CVPR}, we thus propose a direction-aware neural texture blending to eliminate such artifacts, as shown in Fig. \ref{fig:pipeline_net}.
The direction and angle between the two source views and target view will be an important cue for neural rendering quality, especially under occluded scenarios. 
Given novel view depth $D_{n}$ and source view depth $D_{1}$, $D_{2}$, we wrap them to the novel view $D_{1}^{n}$ and $D_{2}^{n}$, then compute the occlusion map $O_{i} = D_{n}- D_{i}^{n} (i=1,2)$.
Then, we unproject $D_{i}$ to point-clouds.
For each point $P$, we compute the cosine value between $\overrightarrow{c_{i}P}$ and $\overrightarrow{c_{n}P}$ to get angle map $A_{i}$, where $c_{i}$ denotes the optical center of source camera $i$, $c_{n}$ denotes the optical center of novel view camera.
Thus, we introduce a direction-aware blending network $\Theta_{DAN}$ to utilize global feature from image and local feature from human geometry to generate the blending weight map $W$, which can be formulated as:
\begin{align}
	W = \Theta_{DAN}(I_{1},O_{1},A_{1},I_{2},O_{2},A_{2}),
\end{align}

\noindent{\textbf{(d) Spatial-temporal Texture Completion.}}
While human-object interaction activities consistently lead to occlusion, the missing texture on human, therefore, leads to severe artifacts for free-viewpoint rendering.
To end this, we propose a spatial-temporal texture completion method to generate a texture-completed proxy in the canonical human space.
We use the temporal and spatial information to complete the missing texture at view $i$ and time $t$ from different times and different views.

Specifically, we first use the non-rigid deformation to register an up-sampled SMPL model (41330 vertices) with the captured human meshes.
Then, for each point on the proxy, we find the nearest visible points along with all views and all frames, then assign an interpolation color to this point.
We thus generate a canonical human space with the fused texture.
For the occluded part of human in novel view, we render the texture-completed image and blend it with the neural rendering results in Sec. \ref{sec:neuralBlending} (c).

We utilize a layered human-object rendering strategy to render human-object together with the reconstruction and tracking of object.
For each frame, we render human with our novel neural texture blending while rendering objects through a classical graphics pipeline with color correction matrix (CCM).
To combine human and object rendering results, we utilize the depth buffer from the geometry of our human-object capture.

\noindent{\textbf{Training Strategy.}} To enable our sparse-view neural human performance rendering under human-object interaction, we need to train the direction-aware blending network $\Theta_{DAN}$ properly.

We follow \citet{NeuralHumanFVV2021CVPR} to utilize 1457 scans from the Twindom dataset \cite{Twindom} to train our DAN $\Theta_{DAN}$ properly.
Differently, we randomly place the performers inside the virtual camera views and augment this dataset by randomly placing some objects from ShapeNet dataset~\cite{chang2015shapenet}.
By simulating the occlusion of human-object interaction, we make our network more robust to occluded human.
Our training dataset contains RGB images, depth maps and angle maps for all the views and models.

For the training of our direction-aware blending network $\Theta_{DAN}$, we set out to apply the following learning scheme to enable more robust blending weight learning.
The appearance loss function with the perceptual term ~\cite{Johnson2016Perceptual} is to make the blended texture as close as possible to the ground truth, which is formulated as:
\begin{align}
	\left.\mathcal{L}_{r g b}=\frac{1}{n} \sum_{j}^{n}
	\left(
	\left\|I_{r}^{j}-I_{g t}^{j}\right\|_{2}^{2}
	+\left\|\varphi
	\left(
	I_{r}^{j}
	\right)
	-\varphi
	\left(
	I_{g t}^{j}
	\right)
	\right\|_{2}^{2}
	\right) \right.
\end{align}
where $I_{g t}$ is the ground truth RGB images; $\varphi(\cdot)$ denotes the output features of the third-layer of pre-trained VGG-19.

With the aid of such occlusion analysis, our texturing scheme maps the input adjacent images into a photo-realistic texture output of human-object activities in the target view through efficient blending weight learning, without requiring further per-scene training.

%% file: sections/Experiment.tex
\section{Experiment}
In this section, we report the details of our approach and evaluate our method on a variety of complex human-object interaction scenarios. 
All of our experiments are run on a PC with 2.2GHz Intel Xeon 4214 CPU, 32GB RAM, and Nvidia GeForce TITAN RTX GPUs. 
The inputs of our system are from a multi-camera system with six synchronized RGB cameras. 
Fig.~\ref{fig:gallery} demonstrates that our approach faithfully reconstructs the geometry and texture of both human and object under interactions, and even handles severe occlusion and multi-object scenarios, such as pulling a chair and catching two balls.

\noindent{\textbf{Implementation Details.}} 
We assume the perspective camera model in our pipeline. 
We adopt a U-net architecture following NeuralHumanFVV~\cite{NeuralHumanFVV2021CVPR} for the image encoder and a 3D convolution network like IF-Net~\cite{chibane2020implicit} with fewer feature dimensions at each resolution for the voxel encoder. 
U-net architecture is adopted in our $\Theta_{DAN}$ similar to Siamese Network~\cite{koch2015siamese} which takes the input from two source view separately with shared parameters during the encoding process.
We train our network with 1457 scans from Twindom~\cite{Twindom} augmented with rigging models of different poses and random occlusions.

\subsection{Comparison}
To the best of our knowledge, our approach is the first neural layer-wise free-viewpoint performance rendering approach with human-object interactions using only sparse RGB input.
For thorough comparison, we compare our approach against existing neural rendering methods, including the point-based \textbf{NHR}~\cite{Wu_2020_CVPR}, implicit method \textbf{NeRF}~\cite{nerf} and the hybrid texturing-based \textbf{NHFVV}~\cite{NeuralHumanFVV2021CVPR} using the same training data for a fair comparison.
Note that for the training of \textbf{NHR} we obtain the point-clouds from the depth sensors in our capture system, and we extend the \textbf{NeRF} to dynamic setting using per-frame multi-view RGB images for training.
Besides, the human and object segmented masks are all utilized during the training process of these methods for a fair comparison.

\begin{figure}[t]
	\centering
	\includegraphics[width=\linewidth]{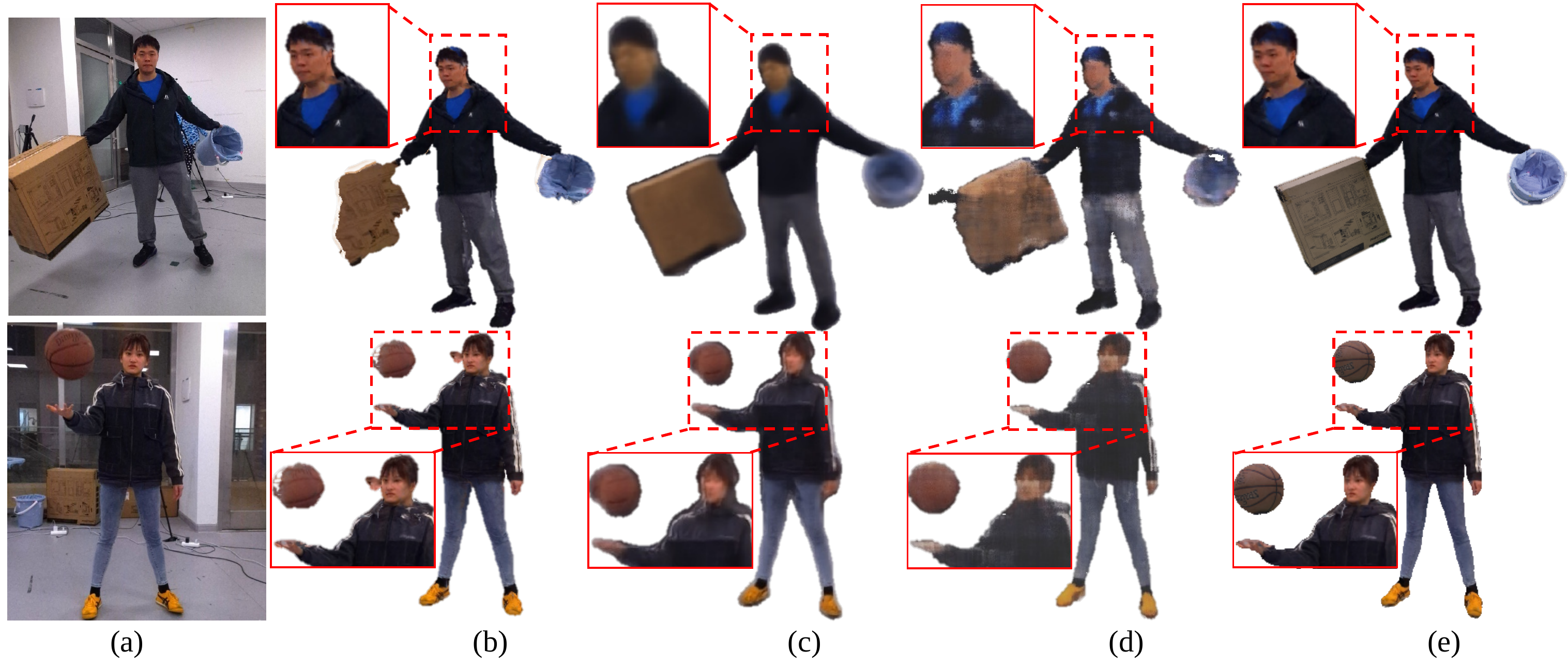}
	\caption{Qualitative comparison on rendering results with various neural rendering approaches. Our approach generates more reasonable and photo-realistic texture. (a) Input images; (b) NHFVV; (c) NHR; (d) NeRF; (e) Ours.}
    \vspace{-10pt}
	\label{fig:comparsion}
\end{figure}

As shown in Fig.~\ref{fig:comparsion}, other method suffers from severe rendering artifacts under our interaction and sparse view setting.
Differently, our approach achieves significantly sharper and more realistic rendering results of human-object interaction scenarios even when serve occlusion appears.
Note that our approach can also enable layer-wise rendering effect and get rid of the tedious per-scene training process.
For quantitative comparison, we adopt the peak signal-to-noise ratio (\textbf{PSNR}), structural similarity index (\textbf{SSIM}), the Mean Absolute Error (\textbf{MAE}) as metrics on the whole testing dataset by comparing the rendering results with source view inputs.
As shown in Tab.~\ref{tab:Comparison}, our approach outperforms other methods on all the metrics above, which illustrates the effectiveness of our approach for free-viewpoint performance rendering under our human-object interaction and sparse-view setting.

\begin{table}[t]
	\begin{center}
		\centering
		\caption{Quantitative comparison against various neural rendering methods on the rendering results. Our method achieves consistently better metric results.}
		\vspace{-10pt}
		\label{tab:Comparison}
		\resizebox{0.4\textwidth}{!}{
			\begin{tabular}{l|cccc}
				\hline
				Method      & PSNR$\uparrow$ & SSIM$\uparrow$ & MAE $\downarrow$  \\
				\hline
				NHFVV~\cite{NeuralHumanFVV2021CVPR}       & 17.545            & 0.966          & 12.949    \\
				NHR~\cite{Wu_2020_CVPR}         & 23.869            & 0.964          & 10.204    \\
				NeRF~\cite{nerf}        & 24.022            & 0.970          & 9.612      \\
				Ours        & \textbf{25.323}   & \textbf{0.985} & \textbf{4.787} \\
				\hline
			\end{tabular}
		}
	\end{center}
\end{table}

\subsection{Evaluation}

\begin{figure}[t]
	\centering
	\includegraphics[width=\linewidth]{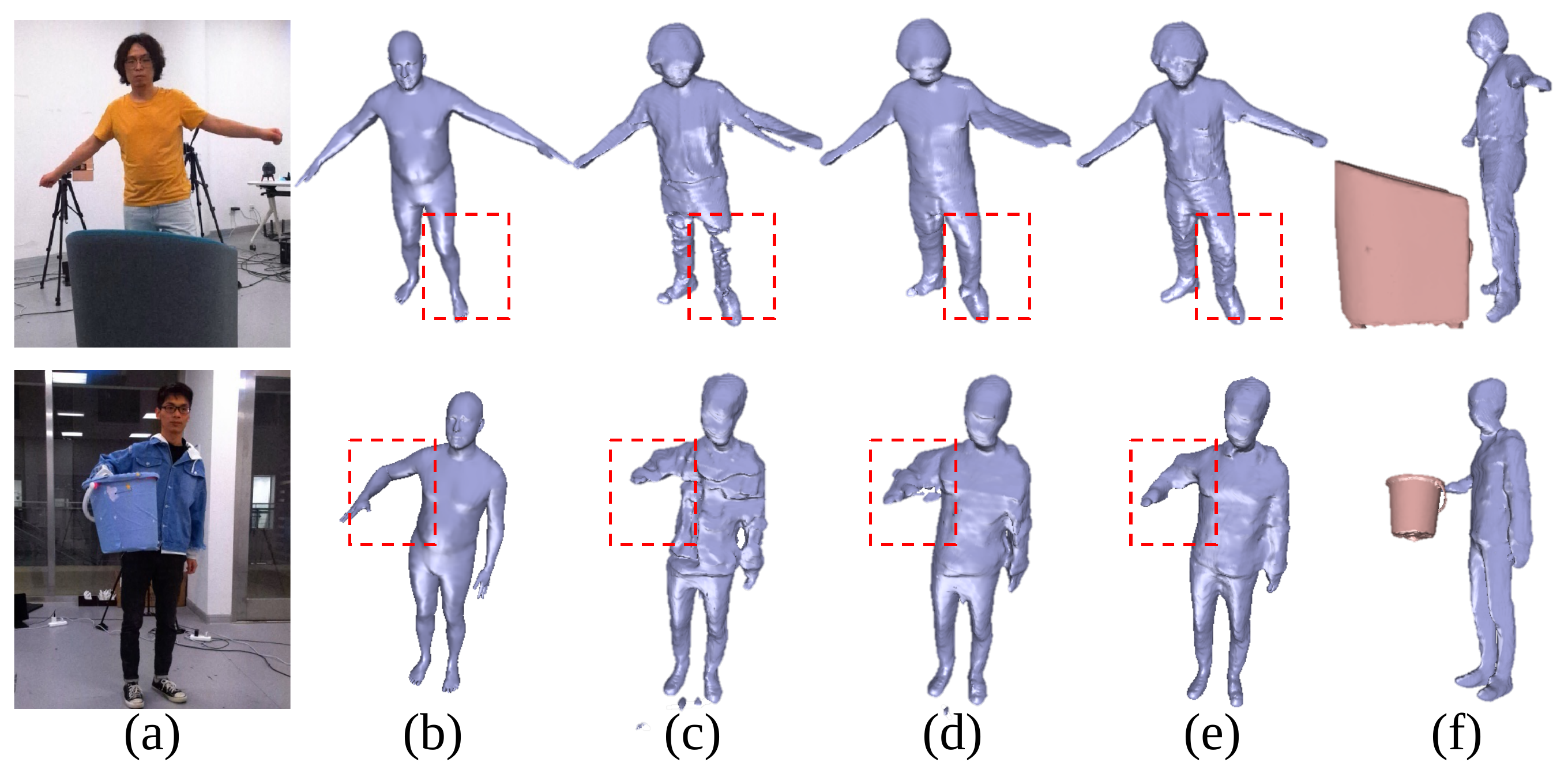}
	\caption{Qualitative evaluation of our occlusion-aware implicit human reconstruction scheme. (a)Input images. (b)w/o implicit; (c)w/o explicit, aug; (d)w/o explicit; (e)ours; (f)ours and object from side view.}
    \vspace{-10pt}
	\label{fig:eval_mesh}
\end{figure}

\noindent{\textbf{Object-aware human reconstruction.}}
Here we evaluate our human geometry reconstruction stage. 
Let \textbf{w/o implicit} denote the variation which only uses explicit parametric human model using the same SMPL fitting process in Sec.~4.1 (a).
Besides, let \textbf{w/o explicit} and let \textbf{w/o explicit, aug} denote the variation without explicit 3D human prior and without both the 3D human prior and occlusion augmentation, respectively.
Fig.~\ref{fig:eval_mesh} provides the qualitative comparison against all these variations in real-world sequences. 
Note that without the implicit geometry inference, only parametric naked human models are recovered without details, while the results without explicit 3D human prior or occlusion augmentation suffer from severe geometry artifacts due to the challenging human-object interactions, especially for the occluded regions and thin structure like arms.
Differently, our approach achieves detailed human geometry reconstruction under challenging occlusions and interactions and can further enable consistent human-object capture through our lay-wise design.
For further quantitative analysis, we evaluate on our synthetic dataset adopt Chamfer distance (\textbf{CD}) in centimeters and Point to Surface distance (\textbf{P2S}) in centimeters, as well as the \textbf{Cosine}) and \textbf{L2} distances upon the re-projection normal as our evaluation metrics 
The corresponding Table.~2 highlights the contribution of each part of our geometry generation component and shows that our full method can produce good geometry under the occlusion scenario. 

\begin{table}[t]
	\begin{center}
		\centering
		\caption{Quantitative evaluation of occlusion-aware human reconstruction scheme on synthetic data. }
		\vspace{-10pt}
		\label{tab:Evaluations}
		\resizebox{\linewidth}{!}{
			\begin{tabular}{l|cc|cc}
				\hline
				\multirow{2}{*}{Method }  & \multicolumn{2}{c|}{Mesh} & \multicolumn{2}{c}{Normal}\\
				\cline{2-5}
				~ & CD$\downarrow$ & P2S$\downarrow$ & Cosine$\downarrow$ & L2$\downarrow$ \\
				\hline
				w/o implicit         & 9.680  &  8.736 & 0.362 & 0.595  \\
				w/o explicit, aug    & 3.551  &  3.698 & 0.215 & 0.443  \\
				w/o explicit         & 3.839  &  6.275 & 0.150 & 0.380  \\
				Ours                 & \textbf{1.819}  & \textbf{2.255} & \textbf{0.134} &\textbf{0.353}  \\
				\hline
			\end{tabular}
		}
		\vspace{-10pt}
	\end{center}
\end{table}

\begin{figure}[t]
	\centering
	\includegraphics[width=\linewidth]{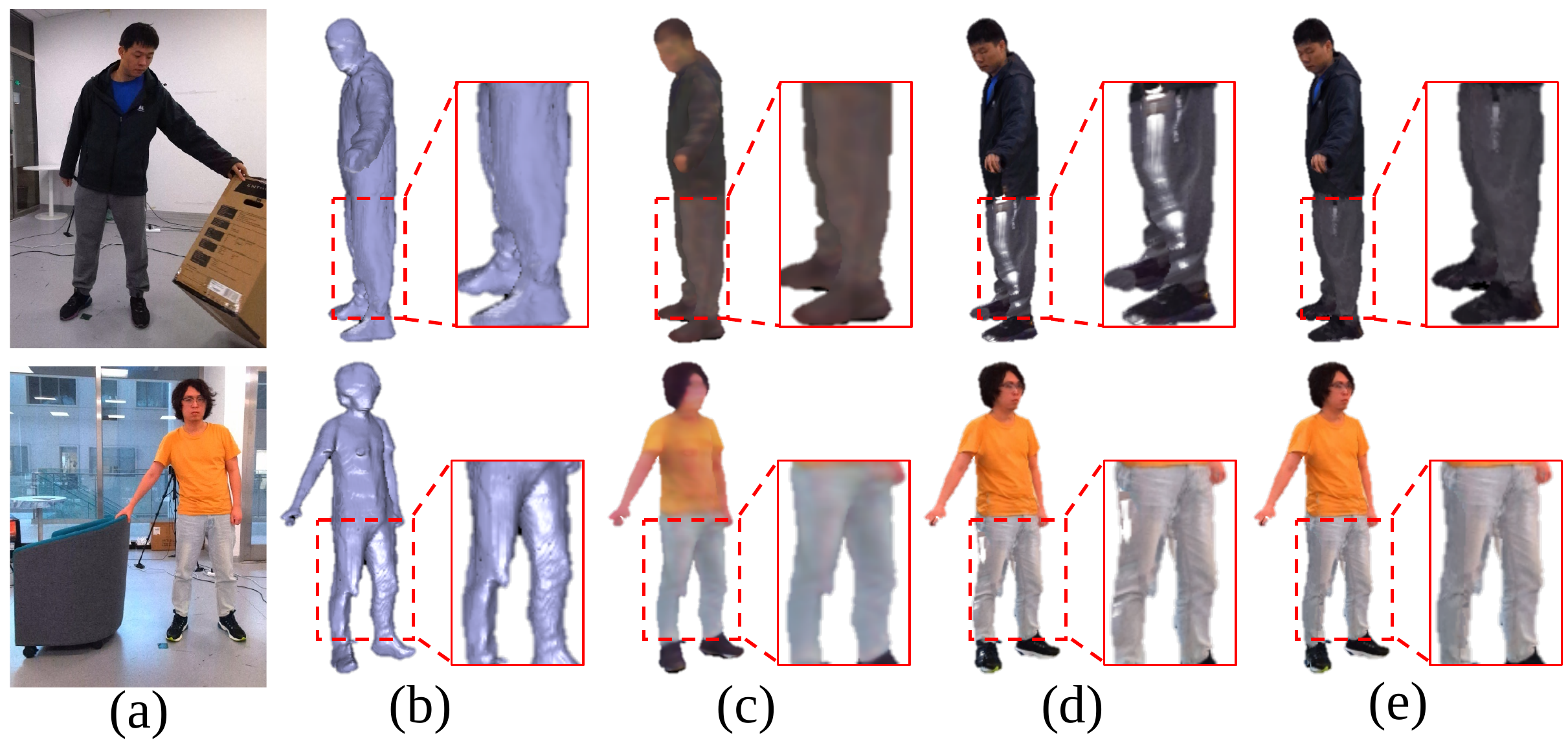}
	\caption{Qualitative evaluation of texturing scheme. (a) Input images; (b) Geometry of our approach; (c) Per-pixel texturing results from PIFu; (d) Neural blending results from NeuralHumanFVV; (e) Our texturing results. }
	\label{fig:eval_rgb}
\end{figure}

\begin{figure}[t]
	\centering
	\includegraphics[width=\linewidth]{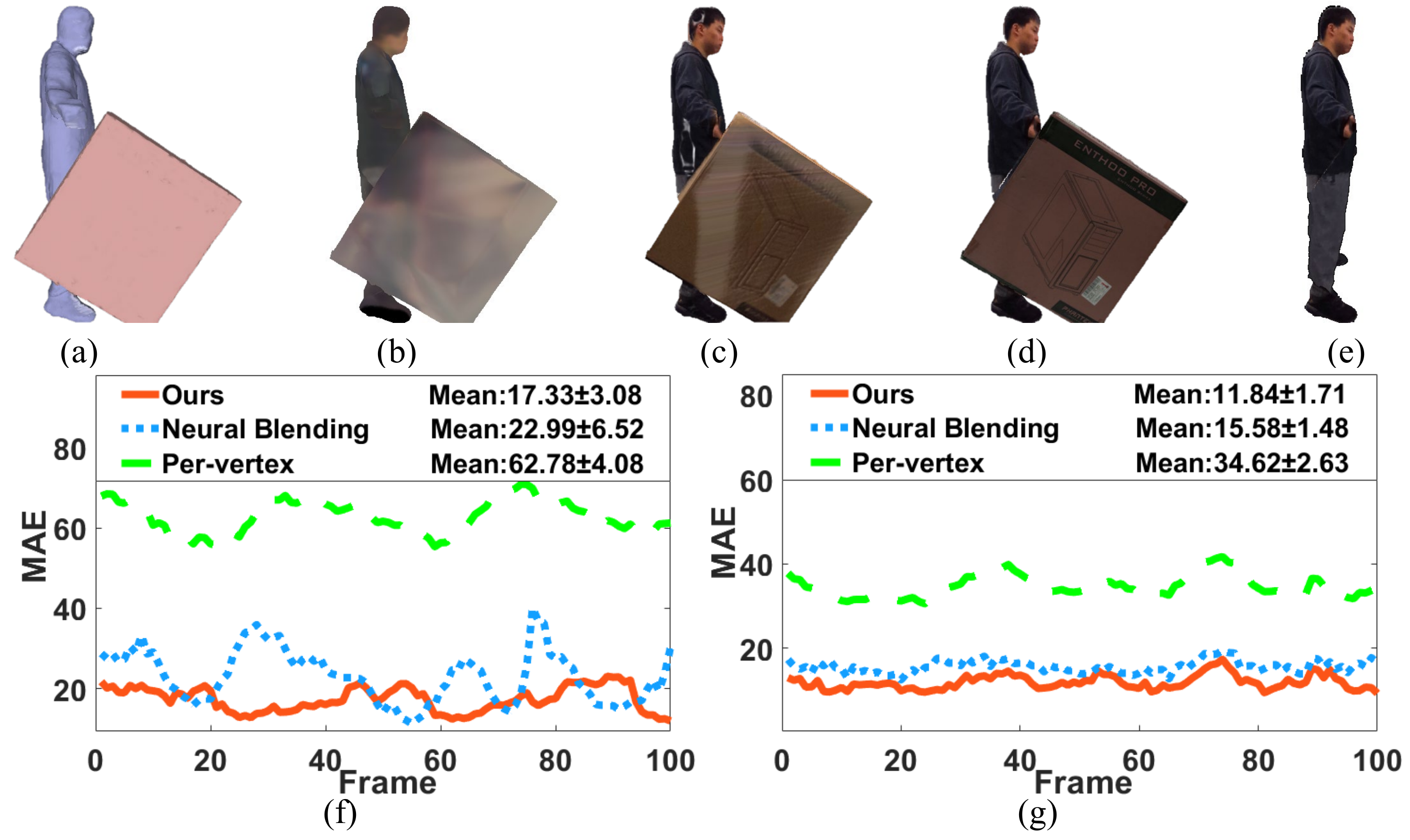}
	\vspace{-20pt}
	\caption{Evaluation of texturing schemes. (a) Our geometry results; (b) Per-pixel texturing results from PIFu; (c) Neural blending results from NHFVV; (d,e) Our texturing results with and w/o object; (f,g) Quantitative results of different texturing schemes with and w/o object, respectively.}
	\label{fig:eval_curve}
	\vspace{-10pt}
\end{figure}

\noindent{\textbf{Layered human-object rendering.}}
We further evaluate our human-object rendering stage against various texturing schemes using the same geometry proxy from our previous stage for a fair evaluation.
Let \textbf{Per-vertex} denote the per-vertex color scheme of PIFu~\cite{PIFU_2019ICCV} and \textbf{Neural Blending} denote the hybrid texturing scheme in NeuralHumanFVV~\cite{NeuralHumanFVV2021CVPR}. 
For qualitative evaluation, as shown in Fig.~\ref{fig:eval_rgb}, both baseline methods suffer from blur texturing results or severe occlusion artifacts near the boundary regions.
In contrast, our layer-wised rendering scheme utilizes both the direction information in an occlusion-aware manner, which generates much sharper and photo-realistic texture rendering results, comparing favorably to the other texturing methods.
Furthermore, we quantitative evaluate against various rendering schemes under the interaction scenarios with objects.
As shown in Fig.~\ref{fig:eval_curve}, the per-vertex scheme suffers from blur texturing artifact while the baseline neural blending scheme suffers from 
occlusion artifacts near the boundary regions especially on the objects.
Differently, our approach generates photo-realistic rendering results under human-object interactions, yielding the lowest mean average error(MAE) of the whole sequence with or without taking the object into error calculation.
These evaluations illustrate the effectiveness of our texturing scheme to utilize the direction information and perform occlusion analysis for complete, photo-realistic, and layer-wise texturing.

\subsection{Limitation and Discussion} 
As a trial for free-viewpoint performance rendering under human-object interactions, the proposed system still owns some limitations.
First, we cannot handle objects with non-rigid deformation or topology changes like tearing a paper, which restricts the application of our systems.
We plan to address it by incorporating the key-volume update technique~\cite{FlyFusion}.
Besides, due to the low image resolution and limited mesh diversity of training data, our method cannot generate good results for fine-grained regions like fingers and handle those extreme poses and occlusions unseen during training. 
A large-scale and high-quality dataset for 4D human-object interaction analysis will be critical for such generalization.
Furthermore, our approach will fail to capture transparent objects which is hard to be segmented.
Our current pipeline models the geometry generation and layer-wise texture rendering separately, and it's an interesting direction to build an end-to-end learning framework such as neural radiance field~\cite{nerf} for complex human-object interaction scenarios.

%% file: sections/Conclusion.tex
\section{Conclusion}
We have presented a neural performance rendering system to generate high-quality geometry and photo-realistic textures of human-object interaction activities in novel views using sparse RGB cameras only. 
Our layer-wise scene decoupling strategy enables explicit disentanglement of human and object for robust reconstruction and photo-realistic rendering under challenging occlusion caused by interactions. 
Specifically, the proposed implicit human-object capture scheme with occlusion-aware human implicit regression and human-aware object tracking enables consistent 4D human-object dynamic geometry reconstruction.
Additionally, our layer-wise human-object rendering scheme encodes the occlusion information and human motion priors to provide high-resolution and photo-realistic texture results of interaction activities in the novel views.
Extensive experimental results demonstrate the effectiveness of our approach for compelling performance capture and rendering in various challenging scenarios with human-object interactions under the sparse setting.
We believe that it is a critical step for dynamic reconstruction under human-object interactions and neural human performance analysis, with many potential applications in VR/AR, entertainment,  human behavior analysis and immersive telepresence.

%% file: sections/Acknowledgments.tex
\section{Acknowledgments}

This work was supported by NSFC programs (61976138, 61977047), the National Key Research and Development Program (2018YFB2100 500), STCSM (2015F0203-000-06), SHMEC (2019-01-07-00-01-E00003) and Shanghai YangFan Program (21YF1429500).